\documentclass[conference]{IEEEtran}
\ifCLASSINFOpdf
\else
\fi

\usepackage{times}
\usepackage{epsfig}
\usepackage{graphicx}
\usepackage{amsmath}
\usepackage{amssymb}
\usepackage{booktabs}
\usepackage{epstopdf}
\DeclareGraphicsExtensions{.eps}
\usepackage{algorithm}%
\usepackage{algpseudocode}%

\usepackage{multirow}
\usepackage{CJK}
\newcommand{\tabincell}[2]{\begin{tabular}{@{}#1@{}}#2
\end{tabular}}
\begin{document}
%
\title{Deep Convolutional Neural Networks for Action Recognition Using Depth Map Sequences}

\author{Pichao Wang$^{\rm 1}$, Wanqing Li$^{\rm 1}$, Zhimin Gao$^{\rm 1}$, Jing Zhang$^{\rm 1}$, Chang Tang$^{\rm 2}$, and Philip Ogunbona$^{\rm 1}$\\
$^{\rm 1}$Advanced Multimedia Research Lab, University of Wollongong, Australia\\
$^{\rm 2}$School of Electronic Information Engineering, Tianjin University, China\\
{\tt\small pw212@uowmail.edu.au, wanqing@uow.edu.au, \{zg126, jz960\}@uowmail.edu.au}\\
{\tt\small tangchang@tju.edu.cn, philipo@uow.edu.au}
}

\maketitle

\begin{abstract}
Recently, deep learning approach has achieved promising results in various fields of computer vision. In this paper, a new framework called Hierarchical Depth Motion Maps (HDMM) + 3 Channel Deep Convolutional Neural Networks (3ConvNets) is proposed for human action recognition using depth map sequences. Firstly, we rotate the original depth data in 3D pointclouds to mimic the rotation of cameras, so that our algorithms can handle view variant cases. Secondly, in order to effectively extract the body shape and motion information, we generate weighted depth motion maps (DMM) at several temporal scales, referred to as Hierarchical Depth Motion Maps (HDMM). Then, three channels of ConvNets are trained on the HDMMs from three projected orthogonal planes separately. The proposed algorithms are evaluated on MSRAction3D, MSRAction3DExt, UTKinect-Action and MSRDailyActivity3D datasets respectively. We also combine the last three datasets into a larger one (called Combined Dataset) and test the proposed method on it. The results show that our approach can achieve state-of-the-art results on the individual datasets and without dramatical performance degradation on the Combined Dataset.
\end{abstract}

\section{Introduction}

Human action recognition has been an active research topic in computer vision due to its wide range of applications, such as smart surveillance and human-computer interactions. In the past decades, research on action recognition mainly focused on recognising actions from conventional RGB videos.

In the previous video-based motion action recognition, most researchers aimed to design hand-crafted features and achieved significant progress. However, in the evaluation conducted by Wang et al. \cite{wang2009evaluation}, one interesting finding is that there is no universally best hand-engineered feature for all datasets.

Recently, the release of the Microsoft Kinect brings up new opportunities in this field. The Kinect device can provide both depth maps and RGB images in real-time at low cost. Depth maps have several advantages compared to traditional color images. For example, depth maps reflect pure geometry and shape cues, which can often be more discriminative than color and texture in many problems including object segmentation and detection. Moreover, depth maps are insensitive to changes in lighting conditions. Based on depth data, many works \cite{Li2010, wang2012mining, Oreifej2013, yangsuper} have been reported with respect to specific feature descriptors to take advantage of the properties of depth maps. However, all of them are based on hand-crafted features, which are shallow high-dimensional descriptions of local or global spatio-temporal information and their performance varies from dataset to dataset. 

Deep Convolutional Neural Networks (ConvNets) have been demonstrated as an effective class of models for understanding image content, offering state-of-the-art results on image recognition, segmentation, detection and retrieval \cite{farabet2013learning, sermanet2014overfeat, Razavian2014CVPRW, agrawal14analyzing}. With the success of ImageNet classification with ConvNets \cite{krizhevsky2012imagenet}, many works take advantage of trained ImageNet models and achieve very promising performance on several tasks, from attributes classification \cite{zhang2014panda} to image representations \cite{oquab2014learning} to semantic segmentation \cite{girshick2014rich}. The key enabling factors behind these successes are techniques for scaling up the networks to millions of parameters and massive labelled datasets that can support the learning process. In this work, we propose to apply ConvNets to depth map sequences for action recognition. An architecture of Hierarchical Depth Motion Maps (HDMM) + 3 Channel Convolutional Neural Network (3ConvNets) is proposed. HDMM is a technique that can transform the problem of action recognition to image classification and artificially enlarge the training data. Specifically, to make our algorithms more robust to viewpoint variations, we directly process the 3D pointclouds and rotate the depth data into different views. To make full use of the additional body shape and motion information from depth sequences, each rotated depth frame is first projected onto three orthogonal Cartesian planes, and then for each projection view, the absolute differences (motion energy) between consecutive and sub-sampled frames are accumulated through an entire depth video sequence. To weight the importances of different motion energy, a weighted factor is used to make the motion energy more important for the recent poses than the past ones. Three HDMMs are constructed after above steps and three ConvNets are trained on the HDMMs. The final classification scores are combined by late fusion of the three ConvNets.

We evaluate our method on the MSRAction3D, MSRAction3DExt, UTKinect-Action and MSRDailyActivity3D datasets individually and achieve results which are better than or comparable to the state-of-the-art. To further verify the robustness of our method, we combine the last three datasets into a single one and test the proposed method on it. The results show that that our approach could achieve consistent performance without much degradation in performance on the combined dataset.

The main contributions of this paper can be summarized as follows. First of all, we propose a new architecture, namely, HDMM + 3ConvNets for depth-based action recognition, which achieves state-of-the-art results on four datasets. Secondly, our method can handle view variant cases for action recognition to some extent due to the simply and directly processing of 3D pointclouds. Lastly, a large dataset is generated by combining the existing ones to evaluate the stability of the proposed method, because the combined dataset contains large variances of within actions, background, viewpoint and number of samples of each action across the three datasets.

The remainder of this paper is organized as follows. Section 2 reviews the related work on deep learning on 2D video action recognition and action recognition using depth sequences. Section 3 describes the proposed architecture. In Section 4, various experimental results and analysis are presented. Conclusion and future work are made in Section 5.

\section{Related Work}
With the recent resurgence of neural networks invoked by Hinton and others \cite{hinton2006fast}, deep neural architectures have been used as an effective solution for extracting high level features from data. There are a number of attempts to apply deep architectures for 2D video recognition. In \cite{taylor2010convolutional}, spatio-temporal features are leaned unsupervised by a Convolutional Restricted Boltzmann Machine (CRBM) and then plugged into a ConvNet for action recognition. In \cite{ji20133d}, 3D convolutional network is used to automatically learn spatio-temporal features directly from raw data. Recently, several ConvNet architectures for action recognition in \cite{karpathy2014large} is compared based on Sport-1M dataset, comprising 1.1 M YouTube videos of sports activities. They find that for a network, operating on individual video frames, performs similarly to the networks whose input is the stack of frames, which indicates that the learned spatio-temporal features do not capture the motion effectively. In \cite{simonyan2014two}, spatial and temporal streams, are proposed for action recognition. Two ConvNets are trained on the two streams and combined by late fusion. The spatial stream is comprised of individual frames while the temporal stream is stacked by optical flow. However, the best results of all above deep learning methods can only match the state-of-the-art results achieved by hand-crafted features. 

For depth-based action recognition, many works have been reported in the past few years. Li et al. \cite{Li2010} sample points from silhouette of a depth image to obtain a bag of 3D points which are clustered to enable recognition. Yang et al. \cite{Yang2012a} stack differences between projected depth maps as DMM and then use HOG to extract the features on the DMM. This method transforms the problem of action recognition from 3D space to 2D space. In \cite{Oreifej2013}, HON4D is proposed, in which surface normal is extended to 4D space and quantized by regular polychorons. Following this method, Yang and Tian \cite{yangsuper} cluster hypersurface normals and form the polynormal which can be used to jointly capture the local motion and geometry information. Super Normal Vector (SNV) is generated by aggregating the low-level polynormals. However, all of these methods are based on carefully hand designed features, which are restricted to specific datasets and applications.  

Our work is inspired by \cite{Yang2012a} and \cite{simonyan2014two}, where we transform the problem of 3D action recognition to 2D image classification in order to take advantage of trained ImageNet models \cite{krizhevsky2012imagenet}.

\section{HDMM + 3ConvNets}
A depth map can be used to capture the 3D structure and shape information. By projecting the difference between depth maps (DMM) onto three orthogonal Cartesian planes can further characterize the motion information of an action \cite{Yang2012a}. To make our method more robust to viewpoint variances, we directly process the 3D pointclouds and rotate the depth data into different views. In order to explore speed invariance and weight the importance of motion energy in time axis, sub-sampled and weighted HDMM is generated from the rotated projected maps. Three deep ConvNets are trained on three projected planes of HDMM. Late fusion is performed by combining the softmax class posteriors from the three nets. The overall framework is illustrated in Figure 1. Our algorithms can be divided into three modules: Rotation in 3D Pointclouds, Hierarchical DMM and Networks Training \& Class Score Fusion. 

\begin{figure*}[!ht]
\begin{center}
{\includegraphics[height = 80mm, width = 180mm]{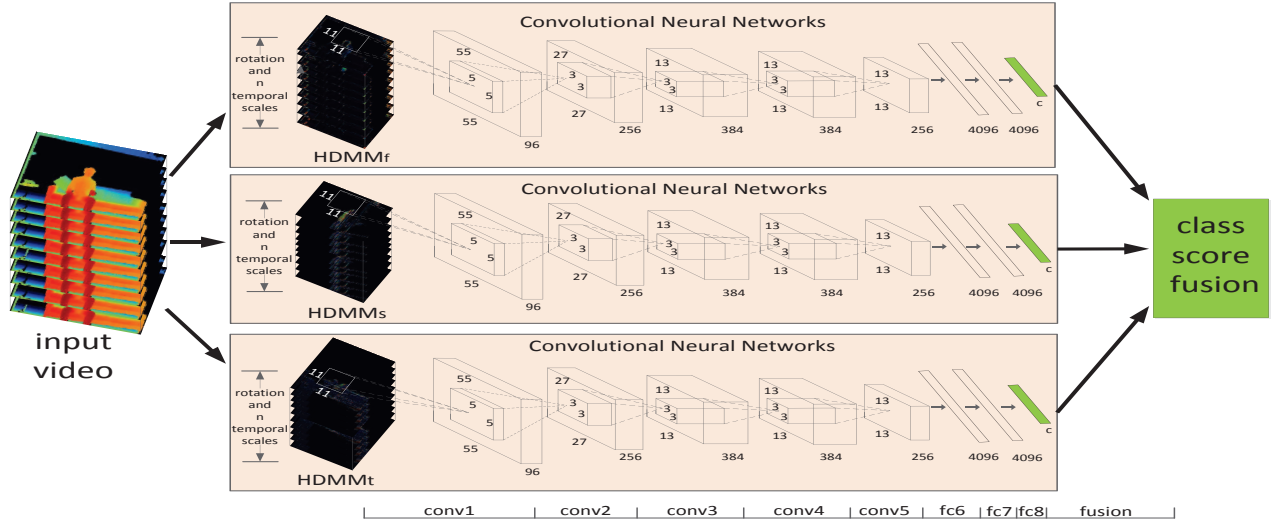}}
\end{center}
\caption{HDMM + 3ConvNets architecture for depth-based action recognition.}
\label{fig:framework}
\end{figure*}

\subsection{Rotation}
One of the challenges for action recognition is the view invariance. To handle this problem, we rotate the depth data in 3D pointclouds, imitating the rotation of cameras around the subject as illustrated in Figure 2 (b), where the rotation is in the world coordinate system (Figure 2 (a)).
\begin{figure}[!ht]
\begin{center}
{\includegraphics[height = 35mm, width = 85mm]{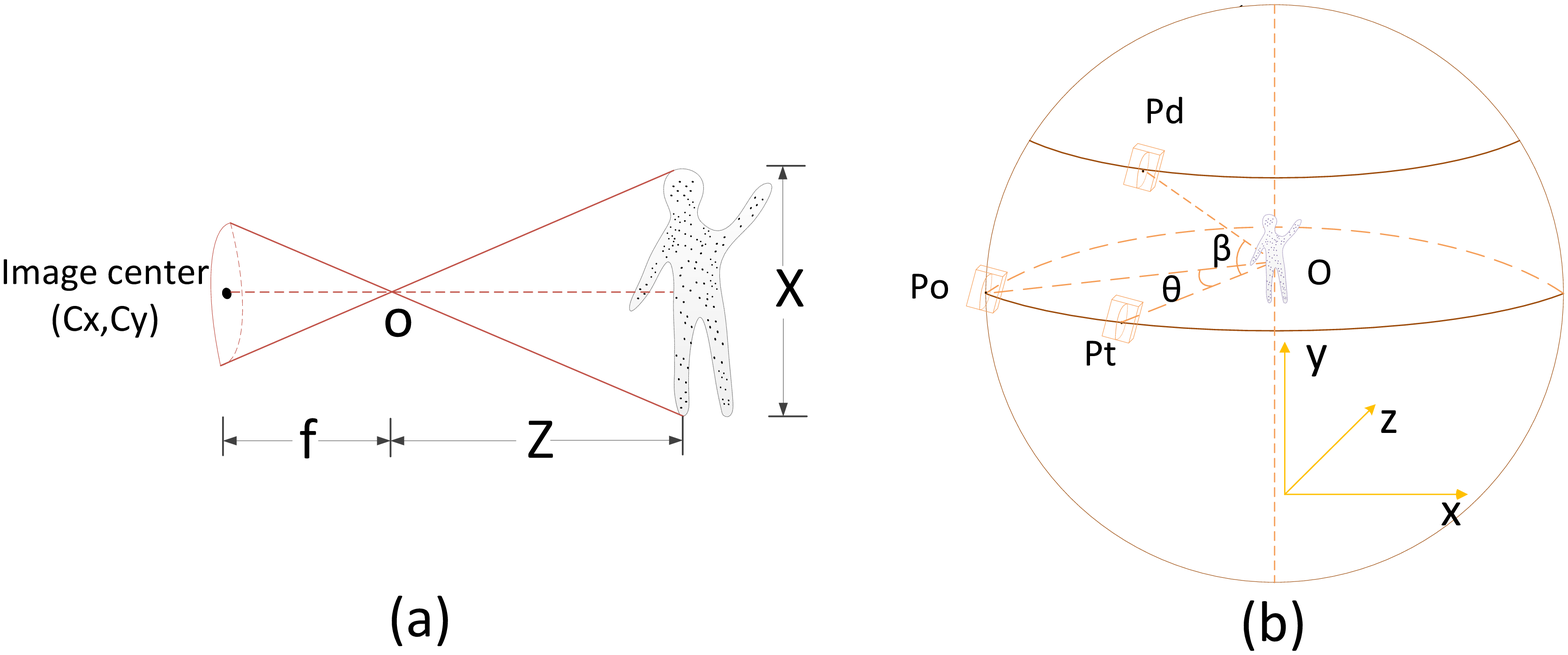}}
\end{center}
\caption{Rotation in 3D Pointclouds.}
\label{fig:framework}
\end{figure}

Figure 2 (b) is the model for rotation of camera around the subject. Supposing camera moves from position $P_{o}$ to $P_{d}$, it can be decomposed into two steps: first moves from $P_{o}$ to $P_{t}$, with rotated angle denoted by $\theta$ and moves from $P_{t}$ to $P_{d}$, with rotated angle denoted by $\beta$. Then the coordinates of subject in rotated scene can be computed by Equation (1).\\
\begin{equation}
\mathbf R  = \mathbf R_{y}\mathbf R_{z}
\begin{bmatrix}
 X, Y, Z, 1 \\     
\end{bmatrix}^ \mathrm{ T }
\end{equation}
where $\mathbf R_{y}$ denotes the rotation around y axis (right-handed coordinate system) while $\mathbf R_{z}$ denotes the rotation around z axis and they are: \\
$\mathbf R_{y} = 
\begin{bmatrix}
 1 & 0 & 0 & 0\\     
 0 & cos(\theta) & -sin(\theta) & Z\cdot sin(\theta)\\ 
 0 & sin(\theta) & cos(\theta) & Z\cdot (1 - cos(\theta))\\
 0 & 0 & 0 & 1\\  
\end{bmatrix}$\\
$\mathbf R_{z} = 
\begin{bmatrix}
 cos(\beta) & 0 & sin(\beta) & -Z\cdot sin(\beta)\\     
 0 & 1 & 0 & 0\\ 
 -sin(\beta) & 0 & cos(\beta) & Z\cdot (1 - cos(\beta))\\
 0 & 0 & 0 & 1\\     
\end{bmatrix} $\\

%
%

After rotation, the 3D cloudpoints are projected to the screen coordinates as illustrated in Figure 2 (a). In this way, the original depth data can be rotated to different angles, with the premise of not resulting in too much information loss.

\subsection{HDMM}

In our work, each rotated 3D depth frame is projected to three orthogonal Cartesian planes, including front, side and top views, denoted by $map_{p}$ where $p \in \{f, s, t\}$. Different from \cite{Yang2012a}, where the motion maps are calculated by accumulating the difference with threshold between consecutive frames, we process the depth maps with three additional steps. Firstly, in order to reserve subtle motion information, for example, page turning when reading books, for each projected map, the motion energy is calculated as the absolute difference between rotated consecutive or sub-sampled frames without thresholding. Secondly, to effectively exploit speed invariance and suppress noise, several temporal scales are generated , as illustrated in Figure 3. 
\begin{figure}[!ht]
\begin{center}
{\includegraphics[height = 40mm, width = 86mm]{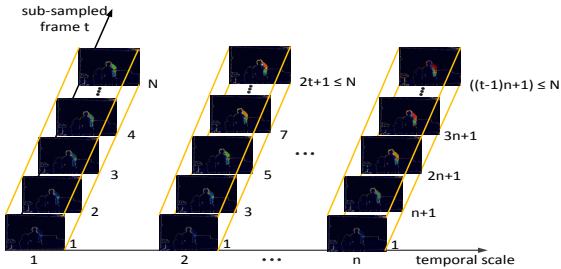}}
\end{center}
\caption{Illustration of hierarchical temporal scales.}
\label{fig:framework}
\end{figure}
For a depth video sequence with $N$ frames, $HDMM_{p}$ is obtained by stacking the motion energy across an entire depth video sequence as follows:
\begin{equation}
HDMM_{p}^{n} = \sum_{t=a}^{b} \left| map_{p}^{(t-1)n + 1} - map_{p}^{(t-2)n + 1} \right|
\end{equation}
where $map_{p}^{i}$ denotes the frame index under projection view $p$ of the whole depth video sequences and $i = (t-1)n + 1$; $t$ represents the sub-sampled frame in corresponding temporal scale $n (n \geq 1)$; $a \in \{2, 3, ..., N\}$ and $\max \limits_{b}\{(b - 1)n + 1 \leq N\}$. Lastly, to weight the different importance of motion energy, a weighted HDMM is adopted as in Equation (3), making the motion energy more important for actions performed currently than past.
\begin{equation}
HDMM_{p}^{t} = \gamma \left| map_{p}^{t} - map_{p}^{t-1} \right | + \delta HDMM_{p}^{t-1}
\end{equation}
Through this simple process, pair actions, such as sit down and stand up, can be differentiated. 

After above three steps, the rotated HDMM are encoded into RGB images, with small values being encoded to R channel while large values to B channel.

\subsection{Network Training \& Class Score Fusion}
After we construct the RGB images from depth motion maps, three ConvNets are trained on the images of the three projected planes. The layer configuration of our three ConvNets is schematically shown in Figure 1, following \cite{krizhevsky2012imagenet}: each net contains eight layers with weights, the first five convolutional layers and the remaining three fully-connected layers. Our implementation is derived from the publicly available Caffe toolbox \cite{jia2014caffe} based on one NVIDIA GeForce GTX680M card.
\subsubsection*{Training}
The training procedure is similar to \cite{krizhevsky2012imagenet}: the network weights are learnt using the mini-batch stochastic gradient descent with momentum set to 0.9 and weight decay set to 0.0005; all hidden weight layers use the rectification (RELU) activation function; at each iteration, a mini-batch of 256 samples is constructed by sampling 256 shuffled training images; all the images are resized to 256 $\times$ 256; to artificially enlarge the training data (data augmentation), firstly, 224 $\times$ 224 patches are randomly cropped from the center of the selected image with a factor of 2048 data augmentation, and then it undergoes random horizontal flipping and RGB jittering; the learning rate is initially set to $10^{-2}$ for directly training the networks from data without initialising the weights with pre-trained models on ILSVRC-2012, while it is set to $10^{-3}$ for fine-tuning with pre-trained models on ILSVRC-2012, and then it is decreased according to a fixed schedule, which is kept the same for all training sets; for each ConvNet we train 100 cycles and decrease the learning rate every 20 cycles. For all experimental settings, we set the dropout regularisation ratio to 0.5 to reduce complex co-adaptations of neurons in nets.

\subsubsection*{Class Score Fusion}
At test period, given a depth video sequence (sample), we only use depth motion maps with temporal scaling but without rotation for testing. The averaged scores of $n$ scales for each test sample are calculated as the final score of this test sample in one channel of 3ConvNets.
The final class scores for a test sample are the averages of the outputs of the three ConvNets. 

\section{Experiments}
In this section, we extensively evaluate our proposed framework on three public benchmark datasets: MSRAction3D \cite{Li2010}, UTKinect-Action \cite{xia2012view} and MSRDailyActivity3D \cite{wang2012mining}. Moreover, an extension of MSRAction3D, called MSRAction3DExt Dataset was used, which contains more subjects performing the same actions. In order to test the stability of proposed method, a new dataset are combined from the last three datasets, referred to as Combined Dataset. In all experiments, for rotation, $\theta$ is set to $(-30^{\circ}:15^{\circ}:30^{\circ})$ and $\beta$ is set to $(-5^{\circ}:5^{\circ}:5^{\circ})$; for weighted HDMM, $\gamma$ is set to 0.99 and $\delta$ is set to 1. Different temporal scales are set according to the noise level and mean circle of actions performed in different datasets. Experimental results show that our method can outperform or match the state-of-the-art on individual datasets and maintain the accuracy on the Combined Dataset.

\subsection{MSRAction3D Dataset}
The MSRAction3D Dataset \cite{Li2010} is an action dataset of depth sequences captured by a depth camera. It contains 20 actions performed by 10 subjects facing the camera, with each subject performing each action 2 or 3 times. The 20 actions are: ``high arm wave", ``horizontal arm wave", ``hammer", ``hand catch", ``forward punch", ``high throw", ``draw X", ``draw tick", ``draw circle", ``hand clap", ``two hand wave", ``side-boxing", ``bend", ``forward kick", ``side kick", ``jogging", ``tennis swing", ``tennis serve", ``golf swing", ``pick up \& throw".

In order to facilitate a fair comparison, the same experimental setting in \cite{wang2012mining} is followed, namely, the cross-subjects settings: subjects 1, 3, 5, 7, 9 for training and subjects 2, 4, 6, 8, 10 for testing. For this dataset, we set temporal scale $n = 1$, and our method achieves 100\% accuracy. Four scenarios are considered: (1) training on primitive data set (without rotation and temporal scaling), (2) training on data set after rotation, (3) pre-training on ILSVRC-2012 (short for pre-trained) followed by fine-tuning on data set after rotation, (4) pre-trained followed by fine-tuning on primitive data set. The results for these setting are listed in Table 1.

\begin{table}[h]
\centering
\caption{Comparison on Different Training Settings for MSRAction3D Dataset.}
\begin{tabular}{|c|c|}
\hline
Training Setting & Accuracy (\%)\\
\hline
Primitive & 7.12\%  \\
\hline
Rotation &34.23\%\\
\hline
Rotation + Pre-trained + Fine-tuning & 100\%  \\
\hline
Primitive + Pre-trained + Fine-tuning & 100\%  \\
\hline
\end{tabular}
\end{table} 
From this table, we can see that pre-training on ILSVRC-2012 (initialise the networks with the trained weights for ImageNet) is important, because the volume of training data is so small that it is not enough to train millions of parameters of the deep networks without good initialisation and leads to overfitting. If we directly train the networks from primitive, the performance is slightly better than random guess.
We compare the performance of HDMM + 3ConvNets with other results in Table 2.

\begin{table}[h]
\centering
\caption{Recognition accuracy comparison of our method and previous approaches on MSRAction3D Dataset.}
\begin{tabular}{|c|c|}
\hline
Method & Accuracy (\%)\\
\hline
Bag of 3D Points \cite{Li2010} & 74.70\%  \\
\hline
HOJ3D \cite{xia2012view} & 79.00\%\\
\hline
Actionlet Ensemble \cite{wang2012mining} & 82.22\%  \\
\hline
Depth Motion Maps \cite{Yang2012a} & 88.73\%  \\
\hline
HON4D \cite{wang2012mining} & 88.89\%  \\
\hline
Moving Pose \cite{zanfir2013moving} & 91.70\%  \\
\hline
SNV \cite{yangsuper} & 93.09\%  \\
\hline
Proposed Method & \textbf{100\%}  \\
\hline
\end{tabular}
\end{table}

The proposed method outperforms all of previous approaches, this is probably because (1) In MSRAction3D we can easily segment the subject from background just by thresholding the depth values, making the generated HDMM without much noise ; (2) Pre-trained models can initialise the image-based deep networks well.

\subsection{MSRAction3DExt Dataset}
The MSRAction3DExt Dataset is an extension of MSRAction3D Dataset. It is captured with the same settings, with additional 13 subjects performing the same 20 actions 2 to 4 times. Thus, there are 20 actions, 23 subjects and 1379 video clips.
Similar to MSRAction3D, we also test our method on the same four scenarios and the results are listed in Table 3. For this dataset, we still adopt cross-subjects setting for training and testing, that is odd subjects for training and even subjects for testing.   

\begin{table}[h]
\centering
\caption{Comparison on Different Training Settings for MSRAction3DExt Dataset.}
\begin{tabular}{|c|c|}
\hline
Training Setting & Accuracy (\%)\\
\hline
Primitive & 10.00\%  \\
\hline
Rotation & 53.05\%\\
\hline
Rotation + Pre-trained + Fine-tuning & 100\%  \\
\hline
Primitive + Pre-trained + Fine-tuning & 100\%  \\
\hline
\end{tabular}
\end{table} 
From Table 1 and Table 3 we can see that with the volume of dataset increasing, directly training the Nets from primitive and rotation, the performance will be much better. However, the performance of trained models will still be very poor if pre-trained model on ImageNet is not used for initialization. Our method again achieves 100\% using pre-trained + fine-tuning even though this dataset has more test samples and variations of actions.
From the two sets of experiments, we can conclude that the way using pre-trained + fine-tuning is very suitable for small datasets. 
In the following experiments, we do not train our networks from primitive any more and all of the experiments adopt pre-trained + fine-tuning settings. 

We compare the performance of our method with SNV \cite{yangsuper} in Table 4 and our method can outperform the state-of-the-art result dramatically.

\begin{table}[h]
\centering
\caption{Recognition accuracy comparison of our method and SNV on MSRAction3DExt Dataset.}
\begin{tabular}{|c|c|}
\hline
Method & Accuracy (\%)\\
\hline
SNV \cite{yangsuper} & 90.54\%  \\
\hline
Proposed Method & \textbf{100\%}  \\
\hline
\end{tabular}
\end{table}

\subsection{UTKinect-Action Dataset}
The UTKinect-Action Dataset \cite{xia2012view} is captured using a stationary Kinect sensor. It consists of 10 actions: ``walk", ``sit down", ``stand up", ``pick up", ``carry", ``throw", ``push", ``pull", ``wave" and ``clap hands". There are 10 different subjects and each subject performs each action twice. This dataset is designed to investigate variations in the view point.

For this dataset, we set temporal scale $n = 5$, to exploit more temporal information in actions. The cross-subjects scheme is followed as in \cite{xia2013spatio} which are different from \cite{xia2012view} where more subjects were used for training in each round. We consider three scenarios for this dataset: (1) pre-trained + fine-tuning on primitive data set; (2) pre-trained + fine-tuning on data set after rotation (3) pre-trained + fine-tuning on data set after rotation and temporal scaling. The results are listed in Table 5. 
\begin{table}[h]
\centering
\caption{Comparison on Different Training Settings for UTKinect-Action Dataset.}
\begin{tabular}{|c|c|}
\hline
Training Setting & Accuracy (\%)\\
\hline
Primitive + Pre-trained + Fine-tuning& 82.83\%  \\
\hline
Rotation + Pre-trained + Fine-tuning & 88.89\%\\
\hline
\tabincell{c}{Rotation + Scaling + \\ Pre-trained + Fine-tuning} & 90.91\%  \\
\hline
\end{tabular}
\end{table} 

From Table 5 we can see that after rotation, it can obtain 6\% improvement in terms of accuracy, which shows that the process of rotation in our method can improve the accuracy greatly. The confusion matrix for the final test is demonstrated in Figure 4， and it shows that the most confused actions are \textit{hand clap} and \textit{ wave}, which share similar shapes of depth motion maps.
\begin{figure}[!ht]
\begin{center}
{\includegraphics[height = 38mm, width = 78mm]{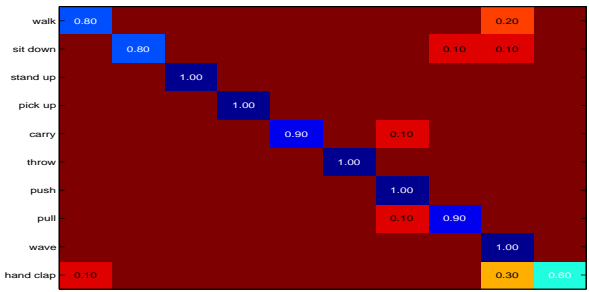}}
\end{center}
\caption{The confusion matrix of proposed method for UTKinect-Action Dataset.}
\label{fig:framework}
\end{figure}

Table 6 shows the performance of our method compared to the previous approaches on the UTKinect-Action Dataset, and it shows that the performance of proposed method can outperform the methods specially designed for view variant cases.

\begin{table}[h]
\centering
\caption{Recognition accuracy comparison of our method and previous approaches on  UTKinect-Action Dataset.}
\begin{tabular}{|c|c|}
\hline
Method & Accuracy (\%)\\
\hline
DSTIP+DCSF \cite{xia2013spatio} & 85.8\%  \\
\hline
Random Forests \cite{zhu2013fusing} & 87.90\%  \\
\hline
SNV \cite{yangsuper} & 88.89\%  \\
\hline
Proposed Method & \textbf{90.91\%}  \\
\hline
\end{tabular}
\end{table}

\subsection{MSRDailyActivity3D Dataset}
The MSRDailyActivity3D Dataset \cite{wang2012mining} is a daily activity dataset of depth sequences captured by a depth camera. There are 16 activities: ``drink", ``eat", ``read book", ``call cellphone", ``write on paper", ``use laptop", ``use vacuum cleaner", ``cheer up", ``sit still", ``toss paper", ``play game", ``lay down on sofa", ``walking", ``play guitar", ``stand up" and ``sit down". There are 10 subjects and each subject performs each activity twice, one in standing position and the other in sitting position. Compared to MSRAction3D(Ext) and UTKinect-Action datasets, actors in this dataset present large spatial and scaling changes. Moreover, most activities in this dataset involve human-object interactions. 

For this dataset, we set temporal scale $n = 21$, a larger number of scales, to exploit more temporal information and suppress the high level noise in this dataset. We follow the same experimental setting as \cite{wang2012mining} and obtain the final accuracy of 81.88\%. Three scenarios are considered for this dataset: (1) pre-trained + fine-tuning on primitive data set; (2) pre-trained + fine-tuning on data set after temporal scaling; (3) pre-trained + fine-tuning on data set after temporal scaling and rotation. The results are listed in Table 7.

\begin{table}[h]
\centering
\caption{Comparison on Different Training Settings for MSRDailyActivity3D Dataset.}
\begin{tabular}{|c|c|}
\hline
Training Setting & Accuracy (\%)\\
\hline
Primitive + Pre-trained + Fine-tuning& 46.25\%  \\
\hline
Scaling + Pre-trained + Fine-tuning & 75.62\%\\
\hline
\tabincell{c}{Scaling + Rotation +\\ Pre-trained + Fine-tuning} & 81.88\%  \\
\hline
\end{tabular}
\end{table} 

The performance of our method compared to the previous approaches is shown in Table 8 and the confusion matrix is shown in Figure 5.

\begin{table}[h]
\centering
\caption{Recognition accuracy comparison of our method and previous approaches on MSRDailyActivity3D Dataset.}
\begin{tabular}{|c|c|}
\hline
Method & Accuracy (\%)\\
\hline
LOP \cite{wang2012mining} & 42.50\%  \\
\hline
Depth Motion Maps \cite{Yang2012a} & 43.13\%  \\
\hline
Joint Position \cite{wang2012mining} & 68.00\%\\
\hline
Moving Pose \cite{zanfir2013moving} & 73.8\%  \\
\hline
Local HON4D \cite{Oreifej2013} & 80.00\%  \\
\hline
Actionlet Ensemble \cite{wang2012mining} & 85.75\%  \\
\hline
SNV \cite{yangsuper} &\textbf{ 86.25\%}  \\
\hline
Proposed Method & 81.88\%  \\
\hline
\end{tabular}
\end{table}

\begin{figure}[!ht]
\begin{center}
{\includegraphics[height = 40mm, width = 80 mm]{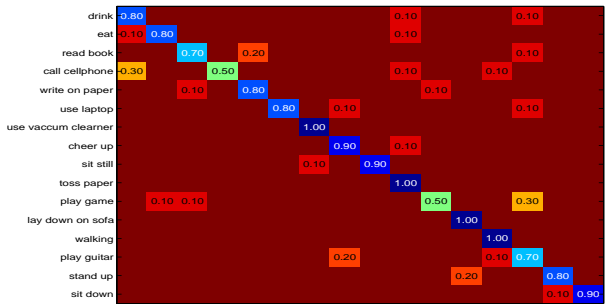}}
\end{center}
\caption{The confusion matrix of proposed method for MSRDailyActivity Dataset.}
\label{fig:framework}
\end{figure}

From Table 8 we can see that our proposed method can outperform DMM \cite{Yang2012a} greatly but can only match the state-of-the-art methods. The reasons probably are: (1) the background of this dataset is more complicated compared to MSRAction3D (Ext), we only pre-process by thresholding the depth value as for MSRAction3D (Ext), which causes lots of noise in HDMM; (2) there are so many actions that are similar, such as \textit{call cellphone}, \textit{drink} and \textit{eat}, they have similar motion shapes but have subtle motion energies so that the HDDM are very similar and confusing.

\subsection{Combined Dataset}
The Combined Dataset is a dataset consisting of MSRAction3DExt, UTKinect-Action and MSRDailyActivity3D datasets to test the scalability of the proposed methods. The Combined Dataset is very challenging due to its large variations in backgrounds, within actions, view points and number of samples in each action. The same actions in different datasets are combined into one action and rewritten into same file format (.bin file format) and filename format (axxx\_sxxx\_exxx\_depth.bin, where a, s and e represent class ID, subject ID and example ID respectively, xxx represent the corresponding number) in Combined Dataset. The combined actions and corresponding information are listed in Table 9.

\begin{table}[h]
\centering
\caption{Information on Combined Dataset: A stands for MSRAction3DExt Dataset; U stands for UTKinect-Action Dataset; D stands for MSRDailyActivity3D Dataset; the corresponding Action Names are shown in Figure 6 (Y label of the fifth sub-figure).}

\begin{tabular}{|c|c|c|c|c|c|}
\hline
 \tabincell{c}{\footnotesize Action\\\footnotesize Label} & \tabincell{c}{\footnotesize Subject\\\footnotesize Label} &  \tabincell{c}{\footnotesize Original\\\footnotesize Datasets} &\tabincell{c}{\footnotesize Action\\\footnotesize Label} &  \tabincell{c}{ \footnotesize Subject\\\footnotesize Label} & \tabincell{c}{\footnotesize Original\\\footnotesize Datasets}\\
\hline
\footnotesize 1 & \footnotesize s001-s023& \footnotesize A & \footnotesize 21 & \footnotesize s001-s020& \footnotesize U\&D\\
\hline
 \footnotesize 2 & \footnotesize s001-s023& \footnotesize A& \footnotesize 22 & \footnotesize s001-s020& \footnotesize U\&D\\
\hline
 \footnotesize 3 & \footnotesize s001-s023& \footnotesize A& \footnotesize 23 & \footnotesize s001-s020& \footnotesize U\&D\\
\hline
\footnotesize 4 &\footnotesize s001-s023&\footnotesize A&\footnotesize 24 &\footnotesize s001-s010&\footnotesize U\\
\hline
\footnotesize 5 &\footnotesize s001-s023&\footnotesize A&\footnotesize 25 &\footnotesize s001-s010&\footnotesize U\\
\hline
\footnotesize 6 &\tabincell{c}{\footnotesize s001-s023\\\footnotesize s025-s034} & \footnotesize A\&U& \footnotesize 26 &\footnotesize  001-s010&\footnotesize U\\
\hline
\footnotesize 7 &\footnotesize s001-s023&\footnotesize A &\footnotesize 27 &\footnotesize s001-s010&\footnotesize U\\
\hline
\footnotesize 8 &\footnotesize s001-s023&\footnotesize A& \footnotesize28 &\footnotesize s001-s010&\footnotesize D\\
\hline
\footnotesize 9 &\footnotesize s001-s023&\footnotesize A&\footnotesize 29&\footnotesize s001-s010&\footnotesize D\\
\hline
\footnotesize 10 &\tabincell{c}{\footnotesize s001-s023\\\footnotesize s025-s034} &\footnotesize A\&U &\footnotesize 30 &\footnotesize s001-s010&\footnotesize D\\
\hline
\footnotesize 11 &\tabincell{c}{\footnotesize s001-s023\\\footnotesize s025-s034} &\footnotesize A\&U&\footnotesize 31 &\footnotesize s001-s010&\footnotesize D\\
\hline
\footnotesize 12 &\footnotesize s001-s023&\footnotesize A&\footnotesize 32 &\footnotesize s001-s010&\footnotesize D\\
\hline
\footnotesize 13 &\footnotesize s001-s023&\footnotesize A&\footnotesize 33 &\footnotesize s001-s010&\footnotesize D\\
\hline
\footnotesize 14 &\footnotesize s001-s023&\footnotesize A&\footnotesize 34 &\footnotesize s001-s010&\footnotesize D\\
\hline
\footnotesize 15 &\footnotesize s001-s023&\footnotesize A &\footnotesize 35&\footnotesize s001-s010&\footnotesize D\\
\hline
\footnotesize 16 &\footnotesize s001-s023&\footnotesize A&\footnotesize 36&\footnotesize s001-s010&\footnotesize D\\
\hline
\footnotesize 17 &\footnotesize s001-s023&\footnotesize A&\footnotesize 37 &\footnotesize s001-s010&\footnotesize D\\
\hline
\footnotesize 18 &\footnotesize s001-s023&\footnotesize A&\footnotesize 38 &\footnotesize s001-s010&\footnotesize D\\
\hline
\footnotesize 19 &\footnotesize s001-s023&\footnotesize A &\footnotesize 39 & \footnotesize s001-s010&\footnotesize D\\
\hline
\footnotesize 20 &\footnotesize s001-s023&\footnotesize A &\footnotesize 40 & \footnotesize s001-s010&\footnotesize D\\
\hline
\end{tabular}
\end{table}

For this dataset, we still use cross-subjects scheme: odd subject ID for training and even subject ID for testing, guaranteeing the training and testing subjects in original datasets with the same identities in Combined Dataset.

To compromise between different datasets, we set temporal scale $n = 5$ for this dataset. Our algorithms are tested with two settings: (1) pre-trained + fine-tuning on primitive data set; (2) pre-trained + fine-tuning on data set after temporal scaling and rotation. The corresponding results are listed in Table 10. 

\begin{table}[h]
\centering
\caption{Comparison on Two Training Settings for the Overall Results on Combined Dataset.}
\begin{tabular}{|c|c|}
\hline
Training Setting & Accuracy (\%)\\
\hline
Primitive + Pre-trained + Fine-tuning& 87.20\%  \\
\hline
\tabincell{c}{Rotation + Scaling + \\ Pre-trained + Fine-tuning} & 90.92\%  \\
\hline
\end{tabular}
\end{table}

From Table 10 we can see that for this big dataset, rotation and scaling are less effective compared with that on smaller dataset. One probable reason is that the training of ConvNets can benefit from large primitive data set to fine-tune millions of parameters.
 
We compare our method with SNV \cite{yangsuper} on this dataset: first train one model in the Combined Dataset and then test the model on original datasets and Combined Datasets. The results and corresponding confusion matrix are shown in Table 11 and Figure 6.
\begin{table}[h]
\centering
\caption{Recognition accuracy comparison of SNV and our method on Combined Dataset and its original datasets.}
\begin{tabular}{|c|c|c|} \hline 
\multirow{2}{*}{Dataset} 
& \multicolumn{2}{c|}{Method}\\
 \cline{2-3}   
 & SNV  & Proposed Method \\ \hline
  MSRAction3D & 89.83\%  & \textbf{94.58\%} \\ \hline
  MSRAction3DExt & 91.15\%  & \textbf{94.05\%} \\ \hline
  UTKinect-Action & \textbf{93.94\%}  & 91.92\% \\ \hline
  MSRDailyActivity3D &60.63\%  & \textbf{78.12\%}  \\ \hline
  Combined  & 86.11\% &   \textbf{90.92\%}  \\ \hline
\end{tabular}
\end{table}

\begin{figure}   
\begin{minipage}[t]{0.5\linewidth} 
\centering   
\includegraphics[width=1.5in]{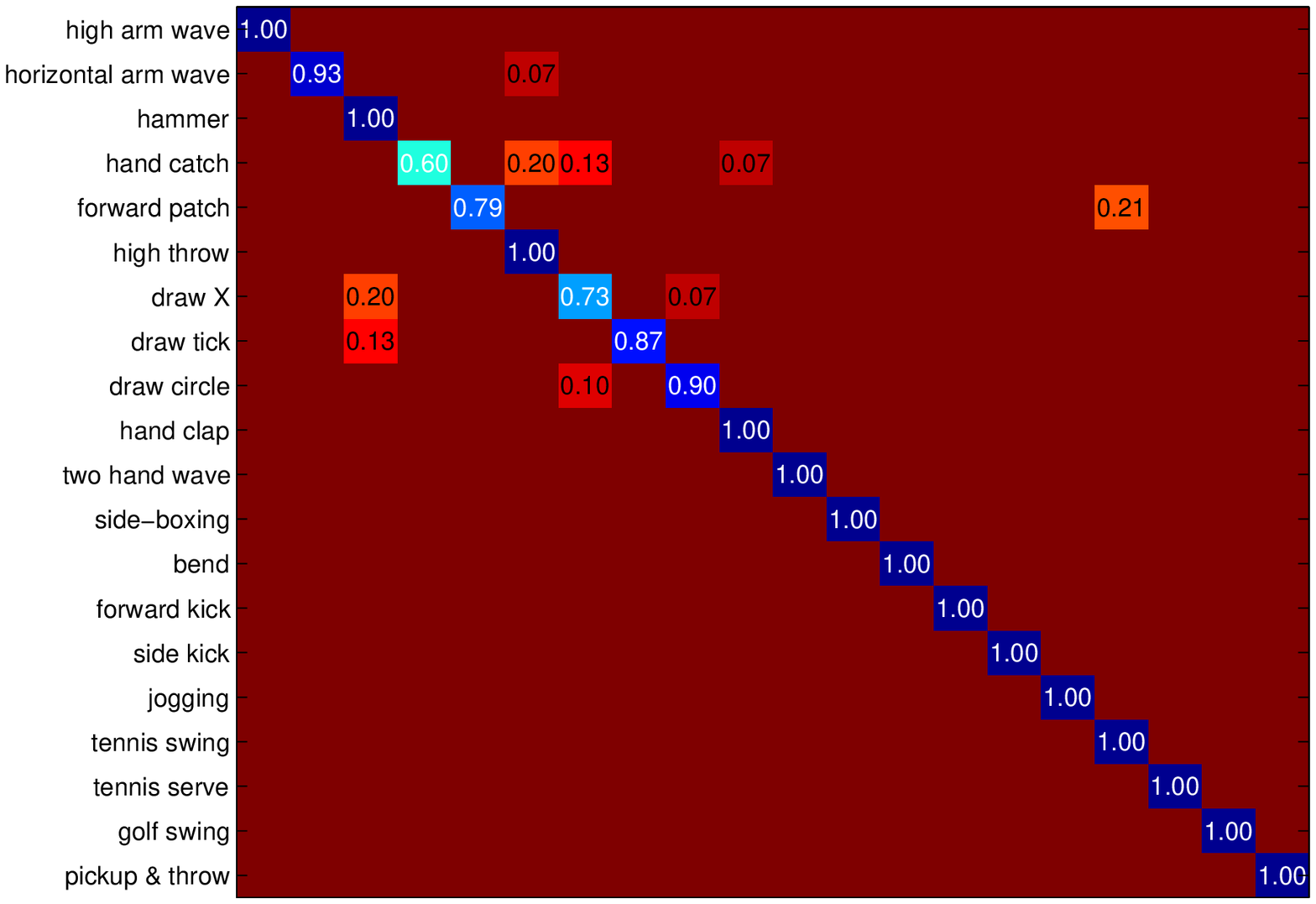}     
\label{fig:side:a}   
\end{minipage}%
\begin{minipage}[t]{0.5\linewidth}   
\centering   
\includegraphics[width=1.5in]{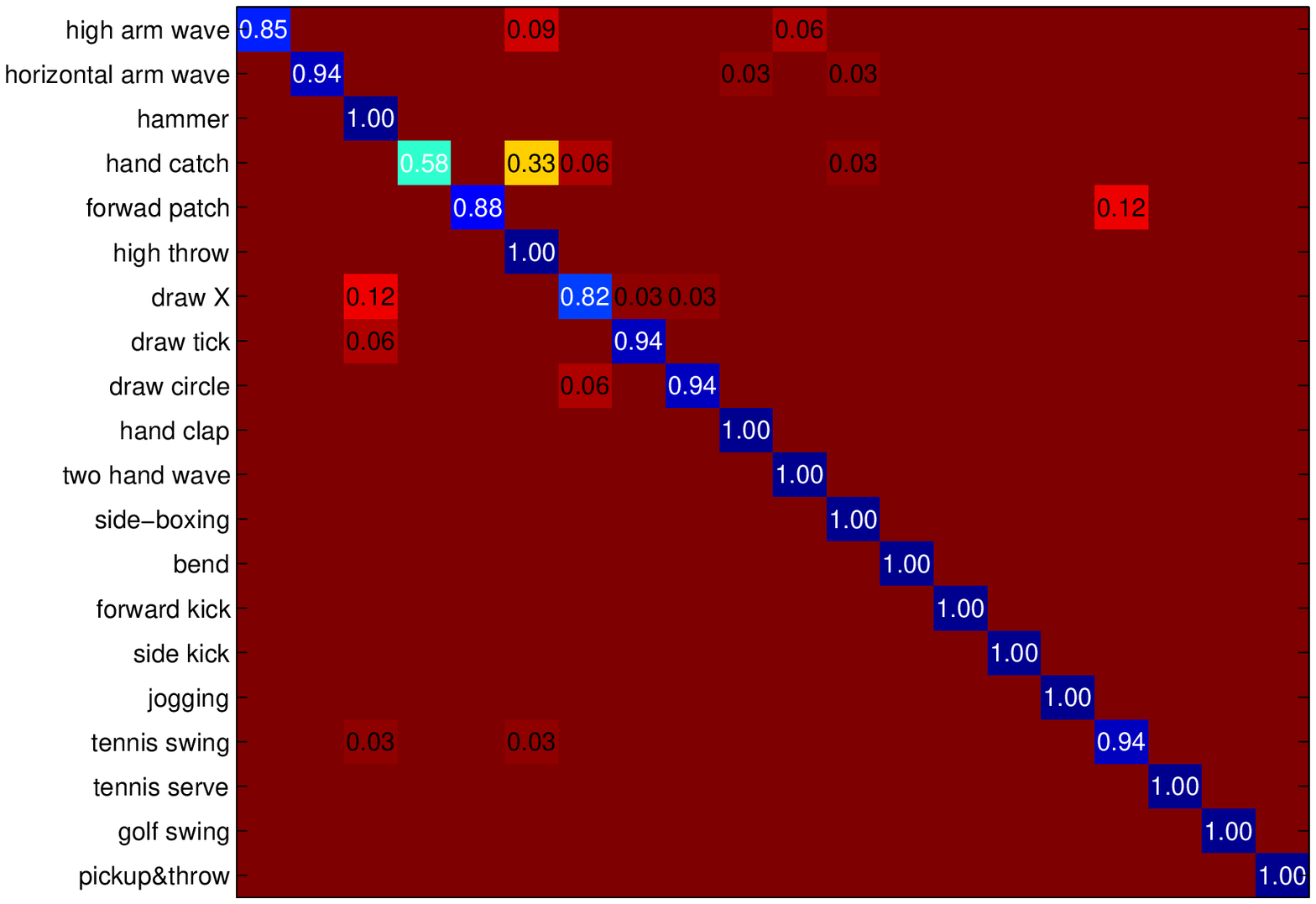}     
\label{fig:side:b}   
\end{minipage}   
\begin{minipage}[t]{0.5\linewidth} 
\centering   
\includegraphics[width=1.5in]{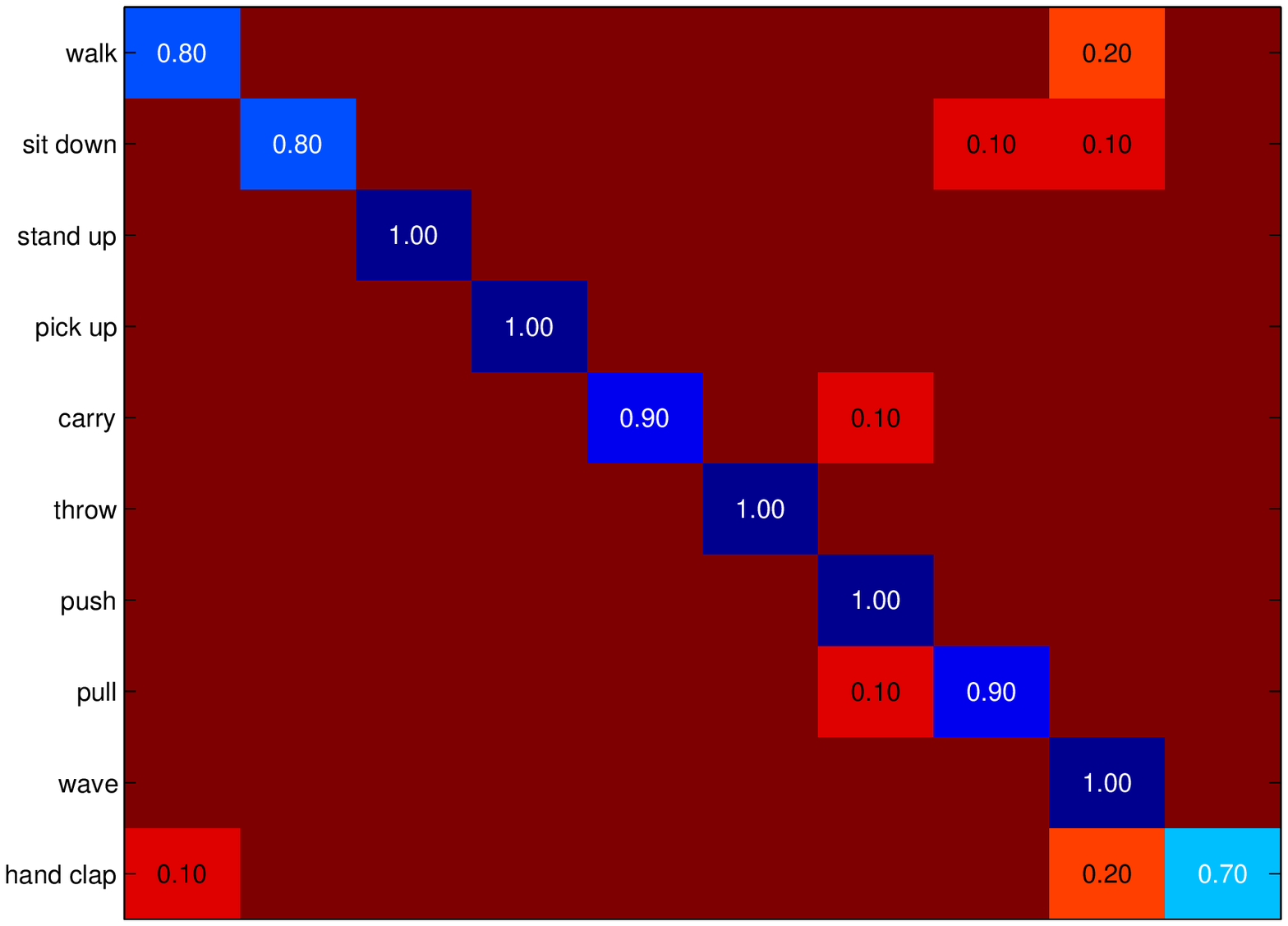}     
\label{fig:side:a}   
\end{minipage}%
\begin{minipage}[t]{0.5\linewidth}   
\centering   
\includegraphics[width=1.5in]{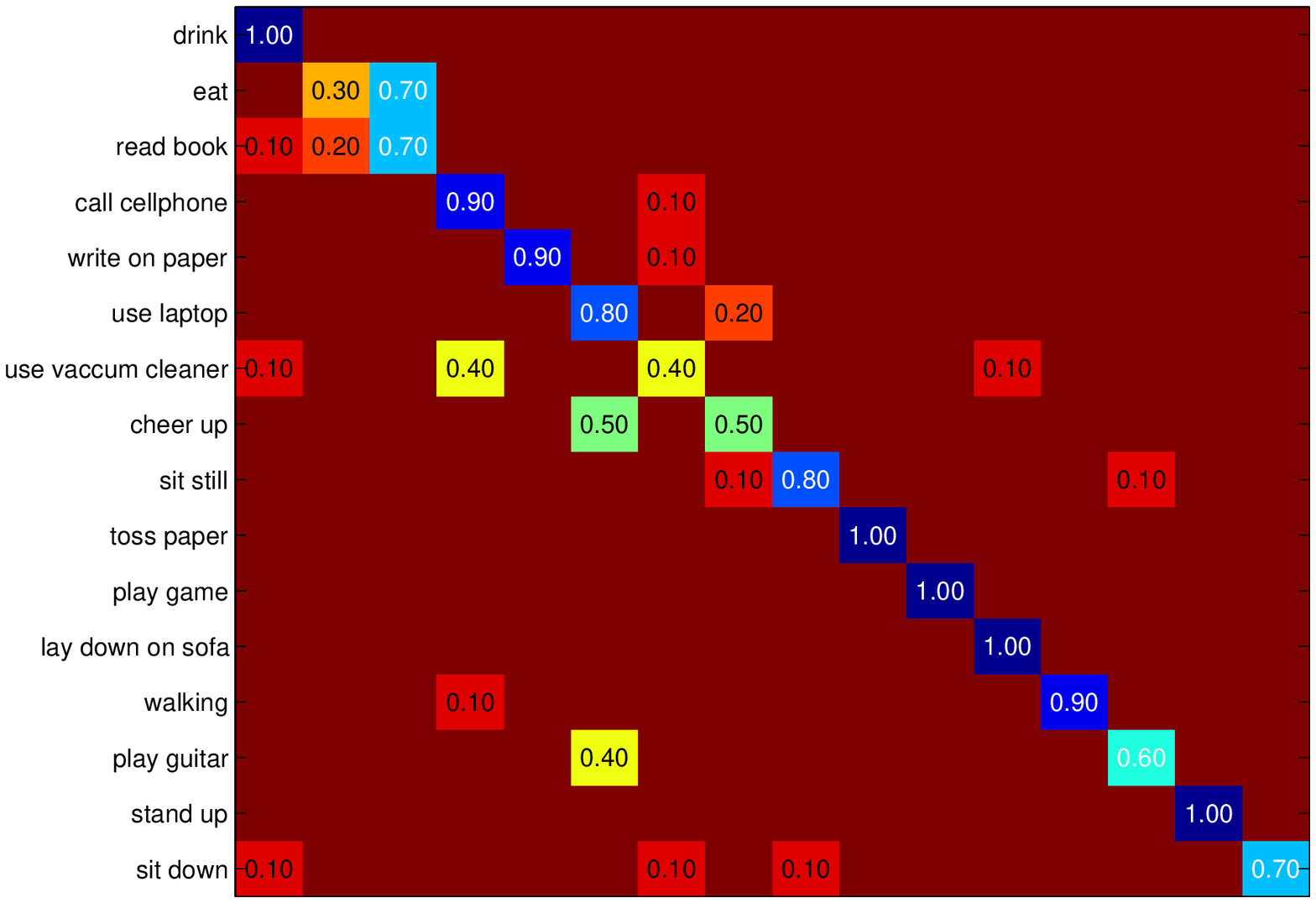}     
\label{fig:side:b}   
\end{minipage} 
\begin{minipage}[t]{1\linewidth}   
\centering   
\includegraphics[width=3.3in]{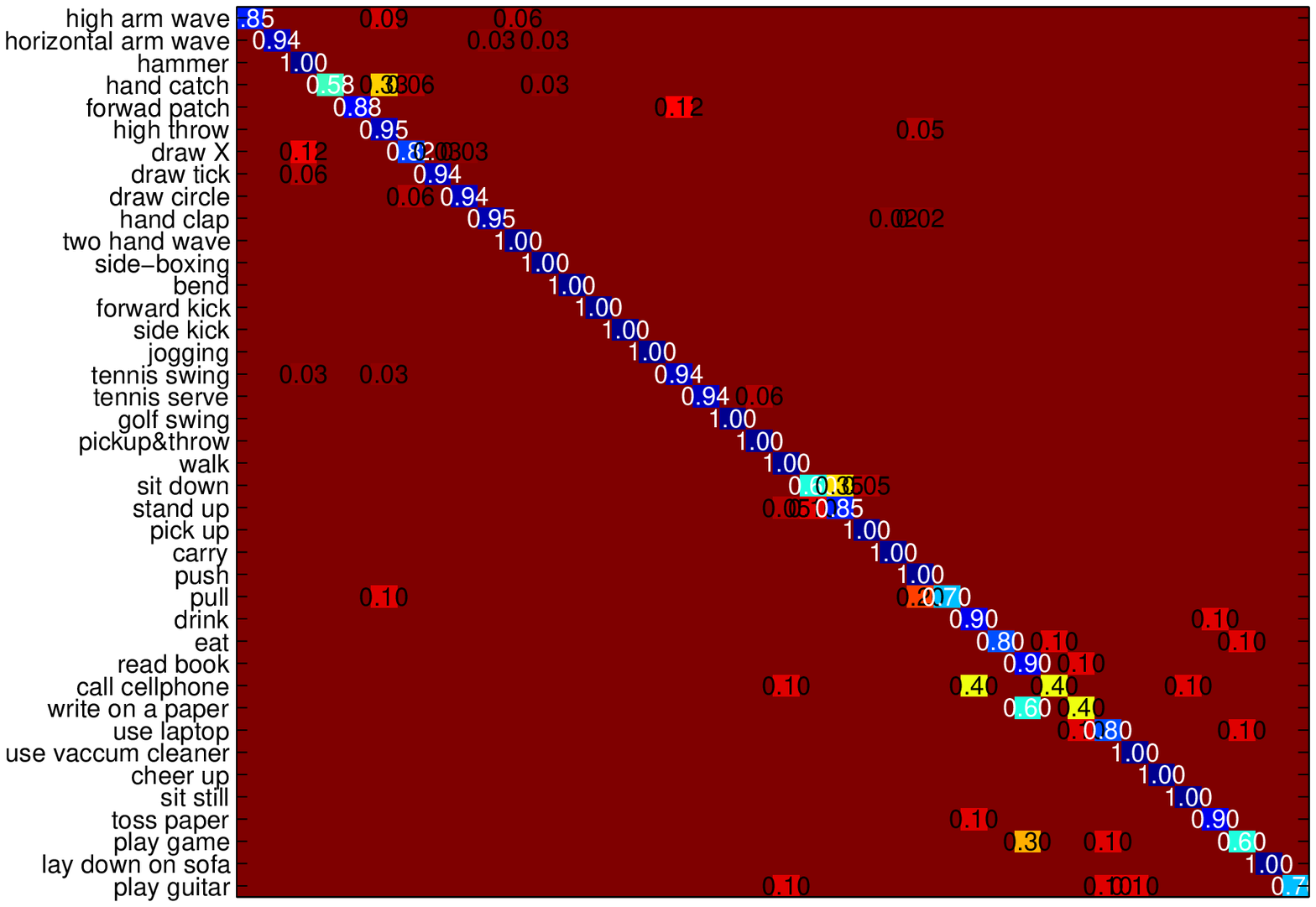}     
\label{fig:side:b}   
\end{minipage} 
\caption{The confusion matrix of proposed method for original MSRAction3D, MSRAction3DExt, UTKinect-Action, MSRDailyActivity and Combined datasets (from top left to down right).}  
\end{figure}  

From Table 11 we can see that proposed method can maintain the accuracy without dramatically variation with the increase of variations and complications of datasets; at the same time, it outperforms the SNV method largely in terms of accuracy in the Combined Dataset. The compromise between different datasets in temporal scales lead to subtle decrease of accuracy in some individual datasets. For example, for MSRDailyActivity3D, decreasing the temporal scale $n = 21$ to $n = 5$ leads to the drop in accuracy, due to the loss of shape and speed information in this dataset, which has high level noise and complicated background. However, for MSRAction3D (Ext), the accuracy drops because some of the actions are similar after temporal scaling, such as \textit{hand catch} and \textit{high throw}, due to the simplicity of these datasets.

\subsection{Analysis}
In this section, we give the overall analysis of proposed methods on data augmentation and parameter selection according to the extensive experimental results. 

In our method, ConvNets are adopted for feature extraction and classification. Generally, ConvNets needs large volume data to tune millions of parameters and reduce overfitting. Directly training the ConvNets with small size of data will lead to very poor performance due to overfitting, which can be seen from Table 1 and Table 3. Due to the small volume (even the Combined Dataset) of available datasets, artificially data augmentation is needed. In our method, two strategies are used for this purpose: rotation and temporal scaling, one for view point invariance, one for speed invariance. However, without initialising the ConvNets with pre-trained model on ImageNet, the artificially enlarged data set are still not enough to train the whole nets from random initialization of the million of weights, because much of the data is interdependent and less informative. The way of pre-trained + fine-tuning provides a promising direction for small datasets in our method. The pre-trained model can initialise the loss function of the nets into minimum areas and the fine-tuning can make the nets obtain the optimum solution even on small datasets.  

For different datasets, different temporal scales are set to obtain the best results. The reasons are as follows: for simple action datasets (or gesture datasets), such as MSRAction3D (Ext), one scale is enough to distinguish the differences between actions (gestures), due to the short circle of motion; for activity datasets, such as UTKinect-Action and MSRDailyActivity datasets, more scales are needed, because the circle of motion in these datasets are much longer and they usually contains several primitive actions (gestures) and large number of scales can capture the motion information in different temporal scales; For noisy datasets, larger temporal scales should be set, such as MSRDailyActivity Dataset, because in order to suppress the affects of high level noise in these datasets, much scales need to be set.

\section{Conclusion and Future Work}
In this paper, we propose a deep classification model for action recognition using depth map sequences. Our proposed method is evaluated on extensive datasets and compared to a number of state-of-the-art approaches. Our method can achieve state-of-the-art results in individual datasets and maintain the accuracy in more complicated datasets combined from available public datasets. By rotation and temporal scaling, the volume of training data can be artificially enlarged, from which the ConvNets benefit and obtain better results than training on primitive. The way of pre-trained + fine-tuning is adopted to train ConvNets on small datasets, which achieves state-of-the-art results in most cases. However, due to the high level noise and complicated background of some datasets, our method can only compete with previous methods. Moreover, our method does not consider the skeleton data on which much success has achieved for action recognition. With the development of deep learning methods, the combination of Deep Belief Networks for skeleton data (generative model) and deep Convolutional Neural Networks for depth data (discriminative model) will open a new door for 3D action recognition. In our future work, we will combine the proposed method together with object segmentation and skeleton-based method to improve the recognition accuracy. 


\bibliographystyle{IEEEtran}
\bibliography{cvpr2015}

\end{document}